\documentclass[11pt,a4paper]{article}
\usepackage[hyperref]{acl2019}
\usepackage{multirow}
\usepackage{latexsym}
\usepackage{mathrsfs}
\usepackage{amsfonts}
\usepackage{amssymb}
\usepackage{amsmath}
\usepackage{bm}
\usepackage{url}
\usepackage{booktabs}
\usepackage{times}
\usepackage{graphicx} 
\usepackage{subfigure}

\newenvironment{itemize*}%
  {\begin{itemize}%
    \setlength{\itemsep}{3pt}%
    \setlength{\parskip}{3pt}}%
  {\end{itemize}}
  \newenvironment{enumerate*}%
  {\begin{enumerate}%
    \setlength{\itemsep}{3pt}%
    \setlength{\parskip}{3pt}}%
  {\end{enumerate}}

\newcommand*\samethanks[1][\value{footnote}]{\footnotemark[#1]}

\def\bs{\mathbf{s}}

\title{Searching for Effective Neural Extractive Summarization: \\ What Works and What's Next}

\aclfinalcopy

\author{Ming Zhong \thanks{\hspace{1mm} These two authors contributed equally.}\\
  \texttt{mzhong18@fudan.edu.cn} \\ \And
  Pengfei Liu \samethanks\\
  \texttt{pfliu14@fudan.edu.cn} \\ \And
  Danqing Wang \\
  \texttt{dqwang18@fudan.edu.cn} \\   \AND
  Xipeng Qiu \thanks{\hspace{1mm} Corresponding author.} \\
  \texttt{xpqiu@fudan.edu.cn} \\ \And
  Xuanjing Huang \\
  \texttt{xjhuang@fudan.edu.cn}
  }

\author{Ming Zhong\thanks{\hspace{1mm} These two authors contributed equally.}, Pengfei Liu\samethanks, Danqing Wang, Xipeng Qiu\thanks{\ \  Corresponding author.} ,  Xuanjing Huang \\
  Shanghai Key Laboratory of Intelligent Information Processing, Fudan University \\
  School of Computer Science, Fudan University \\
  825 Zhangheng Road, Shanghai, China \\
  \texttt{\{mzhong18,pfliu14,dqwang18,xpqiu,xjhuang\}@fudan.edu.cn} }

\begin{document}

\maketitle

\begin{abstract}

    The recent years have seen remarkable success in the use of deep neural networks on text summarization.
    However, there is no clear understanding of \textit{why} they perform so well, or \textit{how} they might be improved.
    In this paper, we seek to better understand how neural extractive summarization systems could benefit from different types of model architectures, transferable knowledge and learning schemas. Additionally, we find an effective way to improve current frameworks and achieve the state-of-the-art result on CNN/DailyMail by a large margin based on our observations and analyses. Hopefully, our work could provide more clues for future research on extractive summarization.
    Source code will be available on Github\footnote{\url{https://github.com/fastnlp/fastNLP}} and our project homepage\footnote{\url{http://pfliu.com/InterpretSum/}}.

\end{abstract}

\section{Introduction}

Recent years has seen remarkable success in the use of deep neural networks for text summarization \cite{see2017get, celikyilmaz2018deep, jadhav2018extractive}.
So far, most research utilizing the neural network for text summarization has revolved around architecture engineering \cite{zhou2018neural, chen2018fast, gehrmann2018bottom}.

Despite their success, it remains poorly understood  why they perform well and what their shortcomings are, which limits our ability to design better architectures.
The rapid development of neural architectures calls for a detailed empirical study of analyzing and understanding existing models.

In this paper, we primarily focus on extractive summarization since they are computationally efficient, and can generate grammatically and coherent summaries \cite{nallapati2017summarunner}.
and seek to better understand how neural network-based approaches to this task could benefit from different types of model architectures, transferable knowledge, and learning schemas, and how they might be improved.

\paragraph{Architectures} Architecturally, the better performance usually comes at the cost of our understanding of the system.
To date, we know little about the functionality of each neural component and the differences between them \cite{peters2018dissecting}, which raises the following typical questions:
    1) How does the choice of different neural architectures (CNN, RNN, Transformer) influence the performance of the summarization system?
    2) Which part of components matters for specific dataset?
    3) Do current models suffer from the over-engineering problem?

Understanding the above questions can not only help us to choose suitable architectures in different application scenarios, but motivate us to move forward to more powerful frameworks.

\paragraph{External Transferable Knowledge and Learning schemas} Clearly, the improvement in accuracy and performance is not merely because of the shift from feature engineering to structure engineering, but the flexible ways to incorporate external knowledge \cite{mikolov2013efficient,peters2018deep,devlin2018bert} and learning schemas to introduce extra instructive constraints \cite{paulus2017deep,arumae2018reinforced}.
For this part, we make some first steps toward answers to the following questions:
    1) Which type of pre-trained models (supervised or unsupervised pre-training) is more friendly to the summarization task?
    2) When architectures are explored exhaustively, can we push the state-of-the-art results to a new level by introducing external transferable knowledge or changing another learning schema?

\begin{table}[htbp]
  \centering \footnotesize
    \begin{tabular}{ll|ll}
    \toprule
    \multicolumn{2}{l}{\textbf{Perspective}} & \textbf{Content} & \multicolumn{1}{l}{\textbf{Sec.ID}} \\
    \midrule
    \multicolumn{2}{l|}{Learning Schemas} & Sup.  \& Reinforce. & \ref{eq:schema} \\
    \midrule
    \multirow{2}[2]{*}{Structure} & Dec.  & Pointer \& SeqLab. & \ref{exp:decoder} \\
          & Enc.  & LSTM \& Transformer & \ref{eq:ecoder} \\
    \midrule
    \multirow{2}[2]{*}{Knowledge} & Exter. & GloVe   BERT  NEWS. & \multirow{2}[2]{*}{\ref{eq:knowledge}} \\
          & Inter. & Random &  \\
    \bottomrule
    \end{tabular}%
      \caption{Outline of our experimental design.
      Dec. and Enc. represent decoder and encoder respectively. Sup. denotes supervised learning and NEWS. means supervised pre-training knowledge.}
  \label{tab:addlabel}%
\end{table}%

To make a comprehensive study of above analytical perspectives, we first build a testbed for summarization system, in which training and testing environment will be constructed.
In the training environment, we design different summarization models to analyze how they influence the performance. Specifically, these models differ in the types of \textbf{architectures} (Encoders: CNN, LSTM, Transformer \cite{vaswani2017attention}; Decoders: auto-regressive\footnote{Auto-regressive indicates that the decoder can make current prediction with knowledge of previous predictions.}, non auto-regressive),  \textbf{external transferable knowledge} (GloVe  \cite{pennington2014glove}, BERT \cite{devlin2018bert}, \textsc{Newsroom} \cite{grusky2018newsroom}) and different \textbf{learning schemas} (supervised learning and reinforcement learning).

To peer into the internal working mechanism of above testing cases, we provide sufficient evaluation scenarios in the testing environment.
Concretely, we present a multi-domain test, sentence shuffling test, and analyze models by different metrics: repetition, sentence length, and position bias,
which we additionally developed to provide a better understanding of the characteristics of different datasets.

Empirically, our main observations are summarized as:

    1) Architecturally speaking, models with auto-regressive decoder are prone to achieving better performance against non auto-regressive decoder. Besides, LSTM is more likely to suffer from the architecture overfitting problem while Transformer is more robust.

    2) The success of extractive summarization system on the CNN/DailyMail corpus heavily relies on the ability to learn positional information of the sentence.

    3) Unsupervised transferable knowledge is more useful than supervised transferable knowledge since the latter one is easily influenced by the domain shift problem.

    4) We find an effective way to improve the current system, and achieving the state-of-the-art result on CNN/DailyMail by a large margin with the help of unsupervised transferable knowledge (\textbf{42.39} R-1 score). And this result can be further enhanced by introducing reinforcement learning (\textbf{42.69} R-1 score).

Hopefully, this detailed empirical study can provide more hints for the follow-up researchers to design better architectures and explore new state-of-the-art results along a right direction.

\section{Related Work}
The work is connected to the following threads of work of NLP research.

\paragraph{Task-oriented Neural Networks Interpreting}

Without knowing the internal working mechanism of the neural network, it is easy for us to get into a hobble when the performance of a task has reached the bottleneck.
 More recently, \citet{peters2018dissecting} investigate how different learning frameworks influence the properties of learned contextualized representations.
 Different from this work, in this paper, we focus on dissecting the neural models for text summarization.

A similar work to us is \citet{kedzie2018content}, which studies how deep learning models perform context selection in terms of several typical summarization architectures, and domains. Compared with this work, we make a more comprehensive study and give more different analytic aspects. For example, we additionally investigate how transferable knowledge influence extractive summarization and a more popular neural architecture, Transformer. Besides, we come to inconsistent conclusions when analyzing the auto-regressive decoder. More importantly, our paper also shows how existing systems can be improved, and we have achieved a state-of-the-art performance on CNN/DailyMail.

\paragraph{Extractive Summarization}
Most of recent work attempt to explore different neural components or their combinations to build an end-to-end learning model. Specifically, these work instantiate their encoder-decoder framework by choosing recurrent neural networks \cite{cheng2016neural, nallapati2017summarunner, zhou2018neural} as encoder, auto-regressive decoder \cite{chen2018fast, jadhav2018extractive, zhou2018neural} or non auto-regressive decoder \cite{isonuma2017extractive, narayan2018ranking, arumae2018reinforced} as decoder, based on pre-trained word representations \cite{mikolov2013efficient, pennington2014glove}. However, how to use Transformer in extractive summarization is still a missing issue. In addition, some work uses reinforcement learning technique \cite{narayan2018ranking, wu2018learning, chen2018fast}, which can provide more direct optimization goals.
Although above work improves the performance of summarization system from different perspectives, yet a comprehensive study remains missing.

\section{A Testbed for Text Summarization}

To analyze neural summarization system, we propose to build a \textit{Training-Testing} environment, in which different text cases (models) are firstly generated under different training settings, and they are further evaluated under different testing settings.
Before the introduction of our Train-Testing testbed, we first give a description of text summarization.

\subsection{Task Description}

Existing methods of extractive summarization directly choose and output the salient sentences (or phrases) in the original document.
Formally, given a document $D = d_1, \cdots, d_n$ consisting of $n$ sentences, the objective is to extract a subset of sentences $R = r_1, \cdots, r_m$  from $D$,
$m$ is deterministic during training while is a hyper-parameter in testing phase.
Additionally, each sentence contains $|d_i|$ words $d_i = x_1,\cdots, x_{|d_i|}$.

Generally, most of existing extractive summarization systems can be abstracted into the following framework, consisting of three major modules: \textbf{sentence encoder}, \textbf{document encoder} and \textbf{decoder}.
At first, a sentence encoder will be utilized to convert each sentence $d_i$ into a sentential representation $\mathbf{d}_i$.
Then these sentence representations will be contextualized by a document encoder to $\mathbf{s}_i$. Finally, a decoder will extract a subset of sentences based on these contextualized sentence representations.

\subsection{Setup for Training Environment}
The objective of this step is to provide typical and diverse testing cases (models) in terms of model architectures, transferable knowledge and learning schemas.

\subsubsection{Sentence Encoder}
We instantiate our sentence encoder with CNN layer \cite{kim2014convolutional}. We don't explore other options as sentence encoder since strong evidence of previous work \cite{kedzie2018content} shows that the differences of existing sentence encoder don't matter too much for final performance.

\subsubsection{Document Encoder}
Given a sequence of sentential representation $\mathbf{d}_1, \cdots, \mathbf{d}_n$, the duty of document encoder is to contextualize each sentence therefore obtaining the contextualized representations $\bs_1, \cdots, \bs_n$.
To achieve this goal,  we investigate the LSTM-based structure and the Transformer structure, both of which have proven to be effective and achieved the state-of-the-art results in many other NLP tasks. Notably, to let the model make the best of its structural bias, stacking deep layers is allowed.

\paragraph{LSTM Layer}
Long short-term memory network (LSTM) was proposed by \cite{hochreiter1997long} to specifically address this issue of learning long-term dependencies, which has proven to be effective in a wide range of NLP tasks, such as text classification \cite{liu2017adversarial,liu2016recurrent}, semantic matching \cite{rocktaschel2015reasoning,liu2016deep}, text summarization \cite{rush2015neural} and machine translation \cite{sutskever2014sequence}.

\paragraph{Transformer Layer}

Transformer \cite{vaswani2017attention} is essentially a feed-forward self-attention architecture, which achieves pairwise interaction by attention mechanism.
Recently, Transformer has achieved great success in many other NLP tasks \cite{vaswani2017attention,dai2018transformer}, and it is appealing to know how this neural module performs on text summarization task.

\subsubsection{Decoder}
Decoder is used to extract a subset of sentences from the original document based on contextualized representations: $\bs_1,\cdots,\bs_n$. Most existing architecture of decoders can divide into auto-regressive and non auto-regressive versions, both of which are investigated in this paper.

\paragraph{Sequence Labeling (SeqLab)}
The models, which formulate extractive summarization task as a sequence labeling problem, are equipped with non auto-regressive decoder.
Formally, given a document $D$ consisting of $n$ sentences $d_1, \cdots, d_n$, the summaries are extracted by predicting a sequence of label $y_1, \cdots, y_n$ ($y_i \in \{0,1\}$) for the document, where $y_i = 1$ represents the $i$-th sentence in the document should be included in the summaries.

\paragraph{Pointer Network (Pointer)}
As a representative of auto-regressive decoder, pointer network-based decoder has shown superior performance for extractive summarization \cite{chen2018fast, jadhav2018extractive}. Pointer network selects the sentence by attention mechanism using  \textit{glimpse} operation \cite{vinyals2015order}. When it extracts a sentence, pointer network is aware of previous predictions.

\subsubsection{External transferable knowledge}
The success of neural network-based models on NLP tasks cannot only be attributed to the shift from feature engineering to structural engineering, but the flexible ways to incorporate external knowledge \cite{mikolov2013efficient,peters2018deep,devlin2018bert}.
The most common form of external transferable knowledge is the parameters pre-trained on other corpora.

To investigate how different pre-trained models influence the summarization system, we take the following pre-trained knowledge into consideration.

\paragraph{Unsupervised transferable knowledge}
Two typical unsupervised transferable knowledge are explored in this paper: context independent word embeddings \cite{mikolov2013efficient,pennington2014glove} and contextualized word embeddings
\cite{peters2018deep,devlin2018bert}, have put the state-of-the-art results to new level on a large number of NLP taks recently.

\paragraph{Supervised pre-trained knowledge}
Besides unsupervised pre-trained knowledge, we also can utilize parameters of networks pre-trained on other summarization datasets.
The value of this investigation is to know transferability between different dataset.
To achieve this, we first pre-train our model on the \textsc{Newsroom} dataset \cite{grusky2018newsroom}, which is one of the largest datasets and contains samples from different domains.
Then, we fine-tune our model on target domains that we investigate.

\subsubsection{Learning Schemas}
Utilizing external knowledge provides a way to seek new state-of-the-art results from the perspective of introducing extra data. Additionally, an alternative way is resorting to change the learning schema of the model. In this paper, we also explore how different learning schemas influence extractive summarization system by comparing supervised learning and reinforcement learning.

\subsection{Setup for Testing Environment}

In the testing environment, we provide sufficient evaluation scenarios to get the internal working mechanism of testing models. Next, we will make a detailed deception.

\paragraph{ROUGE} Following previous work in text summarization, we evaluate the performance of different architectures with the standard ROUGE-1, ROUGE-2 and ROUGE-L {$\rm F_1$} scores \cite{lin2004rouge} by using pyrouge package\footnote{\url{pypi.python.org/pypi/pyrouge/0.1.3}}.

\paragraph{Cross-domain Evaluation}
We present a multi-domain evaluation, in which each testing model will be evaluated on multi-domain datasets based on CNN/DailyMail and \textsc{Newsroom}. Detail of the multi-domain datasets is descried in Tab. \ref{tab:dataset}.

\paragraph{Repetition}
We design repetition score to test how different architectures behave diversely on avoiding generating unnecessary lengthy and repeated information.
We use the percentage of repeated n-grams in extracted summary to measure the word-level repetition, which can be calculated as:
\begin{align}
    \mathrm{REP_n} = \frac{\mathrm{Count} {\mathrm{Uniq}(ngram)}}{\mathrm{Count}(ngram)}
\end{align}
where $\mathrm{Count}$ is used to count the number of n-grams and $\mathrm{Uniq}$ is used to eliminate n-gram duplication. The closer the word-based repetition score is to 1, the lower the repeatability of the words in summary.

\paragraph{Positional Bias} It is meaningful to study whether the ground truth distribution of the datasets is different and how it affects different architectures. To achieve this we design a positional bias to describe the uniformity of ground truth distribution in different datasets, which can be calculated as:
\begin{align}
    \mathrm{PosBias} = \sum_{i=1}^{k}{-{p(i)}\log(p(i))}
\end{align}
We divide each article into $k$ parts (we choose $k=30$ because articles from CNN/DailyMail and \textsc{Newsroom} have 30 sentences by average) and $p(i)$ denotes the probability that the first golden label is in part $i$ of the articles.

\paragraph{Sentence Length} Sentence length will affect different metrics to some extent. We count the average length of the $k$-th sentence extracted from different decoders to explore whether the decoder could perceive the length information of sentences.

\paragraph{Sentence Shuffling} We attempt to explore the impact of sentence position information on different structures. Therefore, we shuffle the orders of sentences  and observe the robustness of different architectures to out-of-order sentences.

\section{Experiment}

\subsection{Datasets}
Instead of evaluating model solely on a single dataset, we care more about how our testing models perform on different types of data,  which allows us to know if current models suffer from the over-engineering problem.

\begin{table}[htbp]
  \centering
    \setlength{\tabcolsep}{2.2mm}{
    \begin{tabular}{lrrr}
    \toprule
    Domains & Train & Valid & Test \\
    \midrule
    CNN/DailyMail & 287,227 & 13,368 & 11,490 \\
    NYTimes & 152,981 & 16,490 & 16,624  \\
    WashingtonPost & 96,775 & 10,103 & 10,196 \\
    FoxNews & 78,795 & 8,428 & 8,397 \\
    TheGuardian & 58,057 & 6,376 & 6,273 \\
    NYDailyNews & 55,653 & 6,057 & 5,904 \\
    WSJ & 49,968 & 5,449 & 5,462 \\
    USAToday & 44,921 & 4,628 & 4,781 \\

    \bottomrule
    \end{tabular}}%
    \caption{Statistics of multi-domain datasets based on CNN/DailyMail and \textsc{NEWSROOM}.}
  \label{tab:dataset}%
\end{table}%

\paragraph{CNN/DailyMail}
The CNN/DailyMail question answering dataset \cite{hermann2015teaching} modified by \cite{nallapati2016abstractive} is commonly used for summarization. The dataset consists of online news articles with paired human-generated summaries (3.75 sentences on average).
For the data prepossessing, we use the data with non-anonymized version as \cite{see2017get}, which doesn't replace named entities.

\paragraph{\textsc{Newsroom}}
Recently, \textsc{Newsroom} is constructed by \cite{grusky2018newsroom}, which contains 1.3 million articles and summaries extracted from 38 major news publications across 20 years. We regard this diversity of sources as a diversity of summarization styles and select seven publications with the largest number of data as different domains to do the cross-domain evaluation. Due to the large scale data in \textsc{Newsroom}, we also choose this dataset to do transfer experiment.

\begin{table*}[t]
\center \footnotesize
\tabcolsep0.07in
\begin{tabular}{llccc|ccc|ccc|ccc}
\toprule
\multicolumn{2}{c}{\textbf{Model}}  &
\textbf{R-1} & \textbf{R-2} & \textbf{R-L} &
\textbf{R-1} & \textbf{R-2} & \textbf{R-L} &
\textbf{R-1} & \textbf{R-2} & \textbf{R-L} &
\textbf{R-1} & \textbf{R-2} & \textbf{R-L} \\
\midrule
\textbf{Dec.} & \textbf{Enc.} &
\multicolumn{3}{c}{\textbf{CNN/DM (2/3)}} &
\multicolumn{3}{c}{\textbf{NYTimes (2)}} &
\multicolumn{3}{c}{\textbf{WashingtonPost (1)}} &
\multicolumn{3}{c}{\textbf{Foxnews (1)}} \\
\cmidrule(lr){1-2} \cmidrule(lr){3-5} \cmidrule(lr){6-8} \cmidrule(lr){9-11} \cmidrule(lr){12-14}
\multicolumn{2}{c}{Lead } & 40.11 & 17.64 & 36.32 & 28.75 & 16.10 & 25.16 & 22.21 & 11.40 & 19.41 & 54.20 & 46.60 & 51.89 \\
\multicolumn{2}{c}{Oracle} & 55.24 & 31.14 & 50.96 & 52.17 & 36.10 & 47.68 & 42.91 & 27.11 & 39.42 & 73.54 & 65.50 & 71.46 \\
\midrule
\multirow{2}{*}{SeqLab}
& LSTM  & 41.22 & 18.72 & 37.52 & 30.26 & 17.18 & 26.58 & 21.27 & 10.78 & 18.56 & 59.32 & 51.82 & 56.95 \\
& Transformer  & 41.31 & \textbf{18.85} & 37.63 & 30.03 & 17.01 & 26.37 & 21.74 & 10.92 & 18.92 & 59.35 & 51.82 & 56.97 \\
\multirow{2}{*}{Pointer}
& LSTM & \textbf{41.56} & 18.77 & \textbf{37.83} & 31.31 & \textbf{17.28} & \textbf{27.23} & \textbf{24.16} & \textbf{11.84} & \textbf{20.67} & \textbf{59.53} & \textbf{51.89} & \textbf{57.08} \\
& Transformer   & 41.36 & 18.59 & 37.67 & \textbf{31.34} & 17.25 & 27.16 & 23.77 & 11.63 & 20.48 & 59.35 & 51.68 & 56.90 \\
\midrule
\textbf{Dec.} & \textbf{Enc.} &
\multicolumn{3}{c}{\textbf{TheGuardian (1)}} &
\multicolumn{3}{c}{\textbf{NYDailyNews (1)}} &
\multicolumn{3}{c}{\textbf{WSJ (1)}} &
\multicolumn{3}{c}{\textbf{USAToday (1)}} \\
\cmidrule(lr){1-2} \cmidrule(lr){3-5} \cmidrule(lr){6-8} \cmidrule(lr){9-11} \cmidrule(lr){12-14}
\multicolumn{2}{c}{Lead} & 22.51 & 7.69 & 17.78 & 45.26 & 35.53 & 42.70 & 39.63 & 27.72 & 36.10 & 29.44 & 18.92 & 26.65 \\
\multicolumn{2}{c}{Oracle} & 41.08 & 21.49 & 35.80 & 73.99 & 64.80 & 72.09 & 57.15 & 43.06 & 53.27 & 47.17 & 33.40 & 44.02 \\
\midrule
\multirow{2}{*}{SeqLab} & LSTM  & 23.02 & 8.12 & 18.29 & 53.13 & 43.52 & 50.53 & 41.94 & 29.54 & 38.19 & 30.30 & 18.96 & 27.40 \\
& Transformer &  23.49 & 8.43 & 18.65 & 53.66 & 44.19 & 51.07 & 42.98 & \textbf{30.22} & 39.02 & 30.97 & 19.77 & 28.03 \\
\multirow{2}{*}{Pointer} & LSTM  & 24.71 & 8.55 & 19.30 & 53.31 & 43.37 & 50.52 & 43.29 & 30.20 & \textbf{39.12} & 31.73 & 19.89 & 28.50 \\
& Transformer & \textbf{24.86} & \textbf{8.66} & \textbf{19.45} & \textbf{54.30} & \textbf{44.70} & \textbf{51.67} & \textbf{43.30} & 30.17 & 39.07 & \textbf{31.95} & \textbf{20.11} & \textbf{28.78} \\

\bottomrule
\end{tabular}
\caption{
Results of different architectures over different domains, where \textbf{Enc.} and \textbf{Dec.} represent document encoder and decoder respectively. Lead means to extract the first $k$ sentences as the summary, usually as a competitive lower bound. Oracle represents the ground truth extracted by the greedy algorithm \cite{nallapati2017summarunner}, usually as the upper bound. The number $k$ in parentheses denotes $k$ sentences are extracted during testing and choose lead-$k$ as a lower bound for this domain. All the experiments use word2vec to obtain word representations.
} \label{tab:all-models}
\end{table*}

\subsection{Training Settings}
For different learning schemas, we utilize cross entropy loss function and reinforcement learning method close to \citet{chen2018fast} with a small difference: we use the precision of ROUGE-1 as a reward for every extracted sentence instead of the $\rm F_1$ value of ROUGE-L.

hird columns show the scope and methods of interactions for different words $w_i$ in a sentence.

For context-independent word representations (GloVe, Word2vec), we directly utilize them to initialize our words of each sentence, which can be fine-tuned during the training phase.

For BERT, we truncate the article to 512 tokens and feed it to a feature-based BERT (without gradient), concatenate the last four layers and get a 128-dimensional token embedding after passing through a MLP.

\subsection{Experimental Observations and Analysis}

Next, we will show our findings and analyses in terms of architectures and external transferable knowledge.

\subsubsection{Analysis of Decoders} \label{exp:decoder}
We understand the differences between decoder \textit{Pointer} and \textit{SeqLab}  by probing their behaviours in different testing environments.

\paragraph{Domains}
From Tab. \ref{tab:all-models}, we can observe that models with pointer-based decoder are prone to achieving better performance against SeqLab-based decoder. Specifically, among these eight datasets, models with pointer-based decoder outperform SeqLab on six domains and achieves comparable results on the other two domains.
For example, in ``\texttt{NYTimes}'', ``\texttt{WashingtonPost}'' and ``\texttt{TheGuardian}'' domains, Pointer surpasses SeqLab by at least $1.0$ improvment (R-1). We attempt to explain this difference from the following three perspectives.

\begin{figure*}[t]
  \centering
  \subfigure[Repetition score]{
    \label{fig:test1}
    \includegraphics[width=0.35\textwidth]{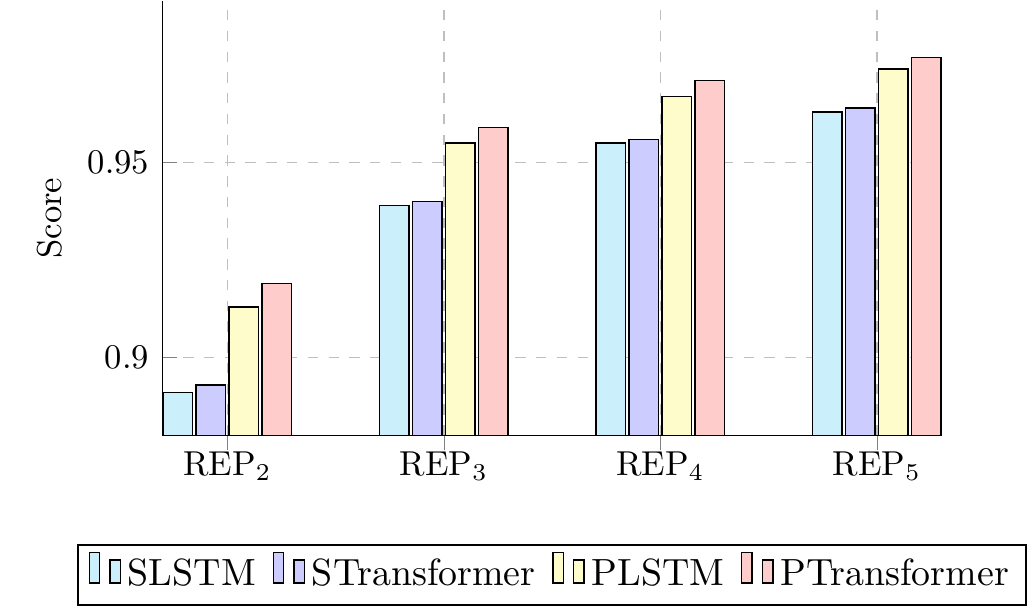}
  }
  \subfigure[Positional bias]{
    \label{fig:test2}
    \includegraphics[width=0.30\textwidth]{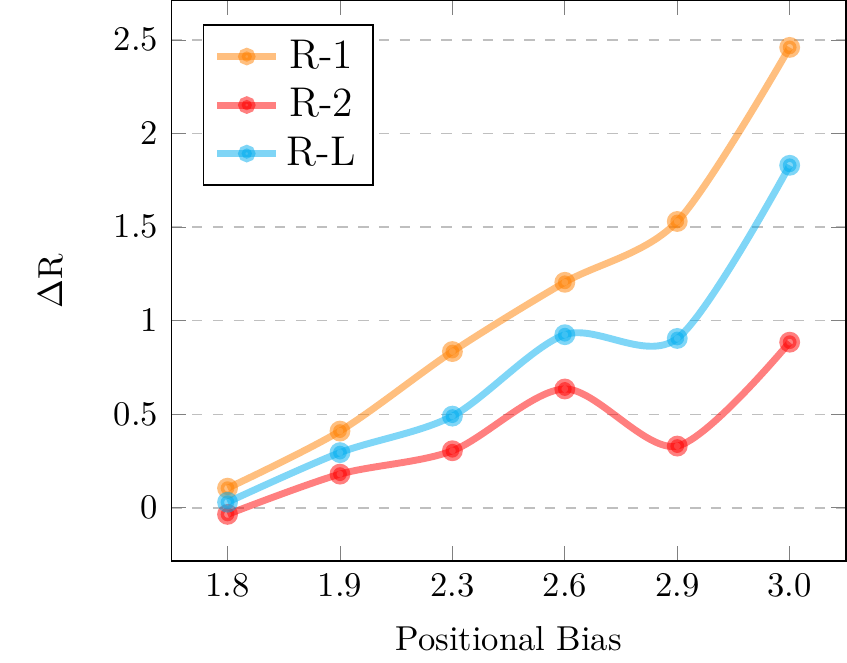}
  }
  \subfigure[Average length]{
    \label{fig:test3}
    \includegraphics[width=0.30\textwidth]{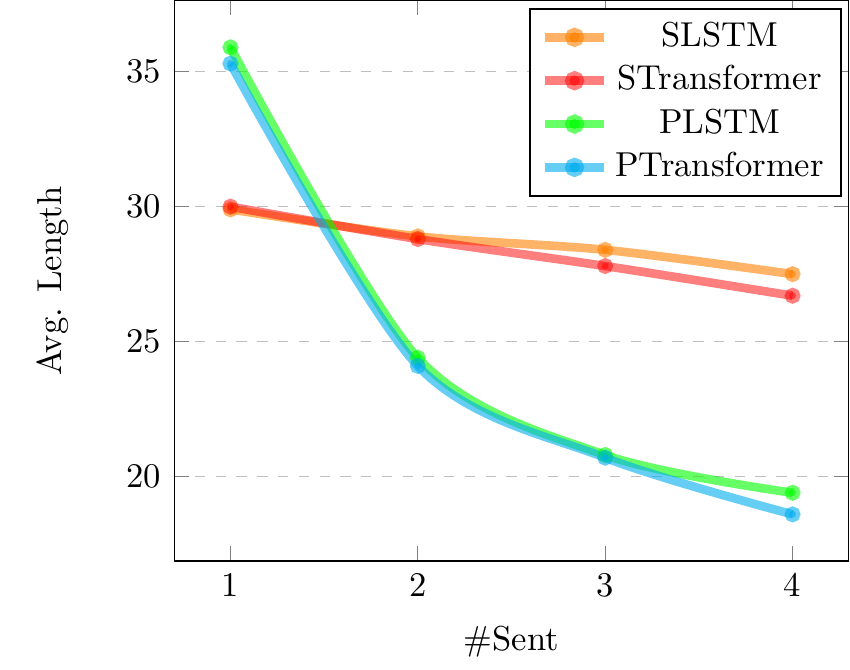}
  }

 \label{fig:three-testing}
 \caption{Different behaviours of two decoders (SeqLab and Pointer) under different testing environment. (a) shows repetition scores of different architectures when extracting six sentences on CNN/DailyMail. (b) shows the relationship between $\Delta \rm{R}$  and positional bias. The abscissa denotes the positional bias of six different datasets and $\Delta \rm{R}$ denotes the average ROUGE difference between the two decoders under different encoders. (c) shows average length of $k$-th sentence extracted from different architectures.}
\end{figure*}

\paragraph{Repetition} For domains that need to extract multiple sentences as the summary (first two domains in Tab. \ref{tab:all-models}), Pointer is aware of the previous prediction which makes it to reduce the duplication of n-grams compared to SeqLab. As shown in Fig. \ref{fig:test1}, models with Pointer always get higher repetition scores than models with SeqLab when extracting six sentences, which indicates that Pointer does capture word-level information from previous selected sentences and has positive effects on subsequent decisions.

\paragraph{Positional Bias} For domains that only need to extract one sentence as the summary (last six domains in Tab. \ref{tab:all-models}), Pointer still performs better than SeqLab. As shown in Fig. \ref{fig:test2}, \textit{the performance gap between these two decoders grows as the positional bias of different datasets increases}. For example, from the Tab. \ref{tab:all-models}, we can see in the domains with low-value positional bias, such as ``\texttt{FoxNews(1.8)}'', ``\texttt{NYDailyNews(1.9)}'', SeqLab achieves closed performance against Pointer. By contrast, the performance gap grows when processing these domains with high-value positional bias (``\texttt{TheGuardian(2.9)}'', ``\texttt{WashingtonPost(3.0)}''). Consequently, SeqLab is more sensitive to positional bias, which impairs its performance on some datasets.

\paragraph{Sentence length} We find \textit{Pointer shows the ability to capture sentence length information based on previous predictions}, while SeqLab doesn't. We can see from the Fig. \ref{fig:test3} that models with Pointer tend to choose longer sentences as the first sentence and greatly reduce the length of the sentence in the subsequent extractions. In comparison, it seems that models with SeqLab tend to extract sentences with similar length.
The ability allows Pointer to adaptively change the length of the extracted sentences, thereby achieving better performance regardless of whether one sentence or multiple sentences are required.

\subsubsection{Analysis of Encoders} \label{eq:ecoder}
In this section, we make the analysis of two encoders LSTM and Transformer in different testing environments.

\paragraph{Domains}
From Tab. \ref{tab:all-models}, we get the following observations:

1)  Transformer can outperform LSTM on some datasets ``\texttt{NYDailyNews}'' by a relatively large margin while LSTM beats Transformer on some domains with closed improvements.
Besides, during different training phases of these eight domains, the hyper-parameters of Transformer keep unchanged\footnote{4 layers 512 dimensions for Pointer and 12 layers 512 dimensions for SeqLab} while for LSTM, many sets of hyper-parameters are used\footnote{the number of layers searches in (2, 4, 6, 8) and dimension searches in (512, 1024, 2048)}.

Above phenomena suggest that LSTM easily suffers from the architecture overfitting problem compared with Transformer.
Additionally, in our experimental setting, Transformer is more efficient to train since it is two or three times faster than LSTM.

2) When equipped with SeqLab decoder, Transformer always obtains a better performance compared with LSTM, the reason we think is due to the non-local bias \cite{wang2018non} of Transformer.

\begin{figure}
    \centering
    \includegraphics[width=0.8\linewidth]{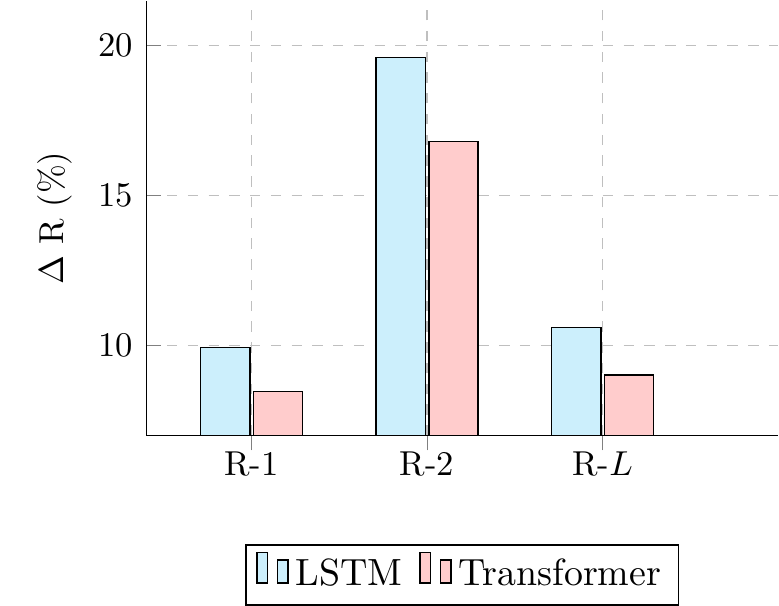}
    \caption{Results of different document encoders with Pointer on normal and shuffled CNN/DailyMail. $\Delta \rm{R}$ denotes the decrease of performance when the sentences in document are shuffled. }
    \label{fig:shuffle}
\end{figure}

\paragraph{Shuffled Testing}

In this settings, we shuffle the orders of sentences in training set while test set keeps unchanged.
We compare two models with different encoders (LSTM, Transformer) and the results can be seen in Fig. \ref{fig:shuffle}.
Generally, there is significant drop of performance about these two models. However, Transformer obtains lower decrease against LSTM, suggesting that Transformer are more robust.

\renewcommand\arraystretch{1.2}
\begin{table}[t]
\center \footnotesize
\tabcolsep0.07in
\setlength{\tabcolsep}{3mm}{
\begin{tabular}{cclll}
\toprule
$\bm{\alpha}$ & $\bm{\beta}$  &
\textbf{R-1} & \textbf{R-2} & \textbf{R-L}  \\
\midrule
1 & 0 & 37.90 & 15.69 & 34.31 \\
$\sqrt{d}$ & 1 & 40.93 & 18.49 & 37.24 \\
1 & 1 & \textbf{41.31} & \textbf{18.85} & \textbf{37.63} \\
1 & $\sqrt{d}$ & 40.88 & 18.42 & 37.19 \\
0 & 1 & 40.39 & 17.67 & 36.54 \\
\midrule

\multicolumn{2}{c}{\citet{nallapati2017summarunner}} & 39.6	& 16.2 & 35.3 \\
\multicolumn{2}{c}{\citet{narayan2018ranking}} & 40.2 & 18.2 & 36.6\\

\bottomrule
\end{tabular}}
\caption{
Results of Transformer with SeqLab using different proportions of sentence embedding and positional embedding on CNN/DailyMail. The input of Transformer is $\alpha$ $*$ sentence embedding plus $\beta$ $*$ positional embedding\protect\footnotemark{}. The bottom half of the table contains models that have similar performance with Transformer that only know positional information.
} \label{tab:disen}
\end{table}
\footnotetext{In \citet{vaswani2017attention}, the input of Transformer is $\sqrt{d}$ $*$ word embedding plus positional embedding, so we design the above different proportions to carry out the disentangling test.}

\renewcommand\arraystretch{1.3}
\begin{table*}[t]
\center \footnotesize
\tabcolsep0.07in
\begin{tabular}{llccc|ccc|ccc|ccc}
\toprule
\multicolumn{2}{c}{\textbf{Model}}  &
\textbf{R-1} & \textbf{R-2} & \textbf{R-L} &
\textbf{R-1} & \textbf{R-2} & \textbf{R-L} &
\textbf{R-1} & \textbf{R-2} & \textbf{R-L} &
\textbf{R-1} & \textbf{R-2} & \textbf{R-L} \\
\midrule
\textbf{Dec.} & \textbf{Enc.} &
\multicolumn{3}{c}{\textbf{Baseline}} &
\multicolumn{3}{c}{\textbf{+ GloVe}} &
\multicolumn{3}{c}{\textbf{+ BERT}} &
\multicolumn{3}{c}{\textbf{\textsc{+ Newsroom}}} \\
\cmidrule(lr){1-2} \cmidrule(lr){3-5} \cmidrule(lr){6-8} \cmidrule(lr){9-11} \cmidrule(lr){12-14}
\multirow{2}{*}{SeqLab}
& LSTM  & 41.22 & 18.72 & 37.52 & \textbf{41.33} & \textbf{18.78} & \textbf{37.64} & 42.18 & 19.64 & 38.53 & 41.48 & \textbf{18.95} & 37.78  \\
& Transformer  & 41.31 & \textbf{18.85} & 37.63 & 40.19 & 18.67 & 37.51 & 42.28 & \textbf{19.73} & 38.59 & 41.32 & 18.83 & 37.63   \\
\multirow{2}{*}{Pointer}
& LSTM & \textbf{41.56} & 18.77 & \textbf{37.83} & 41.15 & 18.38 & 37.43 & \textbf{42.39} & 19.51 & \textbf{38.69} & 41.35 & 18.59 & 37.61  \\
& Transformer   & 41.36 & 18.59 & 37.67 & 41.10 & 18.38 & 37.41 & 42.09 & 19.31 & 38.41 & \textbf{41.54} & 18.73 & \textbf{37.83}  \\

\bottomrule
\end{tabular}
\caption{
Results of different architectures with different pre-trained knowledge on CNN/DailyMail, where \textbf{Enc.} and \textbf{Dec.} represent document encoder and decoder respectively.
} \label{tab:pre-trained}
\end{table*}

\paragraph{Disentangling Testing} 
Transformer provides us an effective way to disentangle position  and content information, which enables us to design a specific experiment, investigating what role positional information plays.

As shown in Tab. \ref{tab:disen}, we dynamically regulate the ratio between sentence embedding and positional embedding by two coefficients $\alpha$ and $\beta$.

\textbf{Surprisingly, we find even only utilizing positional embedding (the model is only told how many sentences the document contains), our model can achieve $\mathbf{40.08}$ on R-1}, which is comparable to many existing models.
By contrast, once the positional information is removed, the performance dropped by a large margin.
This experiment shows that the success of such extractive summarization heavily relies on the ability of learning the positional information on CNN/DailyMail, which has been a benchmark dataset for most of current work.

\subsubsection{Analysis of Transferable Knowledge} \label{eq:knowledge}
Next, we show how different types of transferable knowledge influences our summarization models.

\paragraph{Unsupervised Pre-training}
Here, as a baseline, \textit{word2vec} is used to obtain word representations solely based on the training set of CNN/DailyMail.

As shown in Tab. \ref{tab:pre-trained}, we can find that context-independent word representations can not contribute much to current models.
However, when the models are equipped with BERT, we are excited to observe that the performances of all types of architectures are improved by a large margin.
Specifically, the model  \texttt{CNN-LSTM-Pointer} has achieved a new state-of-the-art with \textbf{42.11} on R-1, surpassing existing models dramatically.

\paragraph{Supervised Pre-training}
In most cases, our models can benefit from the pre-trained parameters learned from the \textsc{newsroom} dataset. However, the model \texttt{CNN-LSTM-Pointer} fails and the performance are decreased.
We understand this phenomenon by the following explanations:
The transferring process from CNN/DailyMail to \textsc{Newsroom} suffers from the domain shift problem, in which the distribution of golden labels' positions are changed. And the observation from Fig. \ref{fig:shuffle} shows that \texttt{CNN-LSTM-Pointer} is more sensitive to the ordering change, therefore obtaining a lower performance.

\paragraph{Why does BERT work?}

We investigate two different ways of using BERT to figure out from where BERT has brought improvement for extractive summarization system.

In the first usage, we feed each individual sentence to BERT to obtain sentence representation, which does not contain contextualized information, and the model gets a high R-1 score of 41.7. However,  when we feed the entire article to BERT to obtain token representations and get the sentence representation through mean pooling, model performance soared to 42.3 R-1 score.

The experiment indicates that though BERT can provide a powerful sentence embedding, the key factor for extractive summarization is contextualized information and this type of information bears the positional relationship between sentences, which has been proven to be critical to extractive summarization task as above.

\renewcommand\arraystretch{1.2}
\begin{table}[t]
\center \footnotesize
\tabcolsep0.07in
\setlength{\tabcolsep}{1mm}{
\setlength{\tabcolsep}{2.5mm}{
\begin{tabular}{llll}
\toprule
{Models} &
\textbf{R-1} & \textbf{R-2} & \textbf{R-L}  \\
\midrule
\citet{chen2018fast} & 41.47 & 18.72 & 37.76 \\
\citet{dong2018banditsum} & 41.50 & 18.70 & 37.60 \\
\citet{zhou2018neural} & 41.59 & 19.01 & 37.98 \\
\citet{jadhav2018extractive}\protect\footnotemark{} & 41.60 & 18.30 & 37.70 \\
\midrule
LSTM + PN                & 41.56   & 18.77   & 37.83 \\
LSTM + PN + RL           & 41.85   & 18.93   & 38.13 \\
LSTM + PN + BERT         & 42.39   & 19.51   & 38.69 \\
LSTM + PN + BERT + RL    &\textbf{42.69}   &\textbf{19.60}   & \textbf{38.85} \\
\bottomrule
\end{tabular}}}
\caption{Evaluation  on CNN/DailyMail. The top half of the table is currently state-of-the-art models, and the lower half is our models.
} \label{table:rl}
\end{table}
\footnotetext{trained and evaluated on the anonymized version.}

\subsection{Learning Schema and Complementarity} \label{eq:schema}
Besides supervised learning, in text summarization, reinforcement learning has been recently used to introduce more constraints.
In this paper, we also explore if several advanced techniques be complementary with each other.

We first choose the based model \texttt{LSTM-Pointer} and \texttt{LSTM-Pointer + BERT}, then the reinforcement learning are introduced aiming to further optimize our models.
As shown in Tab. \ref{table:rl}, we observe that even though the performance of \texttt{LSTM+PN} has been largely improved by BERT, when applying reinforcement learning, the performance can be improved further, which indicates that there is indeed a complementarity between architecture, transferable knowledge and reinforcement learning.

\section{Conclusion}
In this paper, we seek to better understand how neural extractive summarization systems could  benefit  from  different  types  of model architectures, transferable knowledge, and learning  schemas. Our detailed observations can provide more hints for the follow-up researchers to design more powerful learning frameworks.

\section*{Acknowledgment}
We thank Jackie Chi Kit Cheung, Peng Qian for useful comments and discussions.
We would like to thank the anonymous reviewers for their valuable comments. The research work is supported by National Natural Science Foundation of China (No. 61751201 and 61672162),
Shanghai Municipal Science and Technology Commission (16JC1420401 and 17JC1404100),
Shanghai Municipal Science and Technology Major Project(No.2018SHZDZX01)and ZJLab.

\bibliography{./nlp}
\bibliographystyle{acl_natbib}

\end{document}